\documentclass[sigconf, twocolumn]{acmart}
\usepackage{booktabs}
\usepackage{graphicx}
\usepackage{amsmath}
\usepackage{pifont}

\setcopyright{none}

\begin{document}

\title{Semantic Co-Speech Gesture Synthesis and Real-Time Control for Humanoid Robots}

\author{Gang Zhang}
\affiliation{%
  \institution{China Mobile}
  \city{HangZhou}
  \country{China}
}

\begin{abstract}
We present an innovative end-to-end framework for synthesizing semantically meaningful co-speech gestures and deploying them in real-time on a humanoid robot. This system addresses the challenge of creating natural, expressive non-verbal communication for robots by integrating advanced gesture generation techniques with robust physical control. Our core innovation lies in the meticulous integration of a semantics-aware gesture synthesis module, which derives expressive reference motions from speech input by leveraging a generative retrieval mechanism based on large language models (LLMs) and an autoregressive Motion-GPT model. This is coupled with a high-fidelity imitation learning control policy, the MotionTracker, which enables the Unitree G1 humanoid robot to execute these complex motions dynamically and maintain balance. To ensure feasibility, we employ a robust General Motion Retargeting (GMR) method to bridge the embodiment gap between human motion data and the robot platform. Through comprehensive evaluation, we demonstrate that our combined system produces semantically appropriate and rhythmically coherent gestures that are accurately tracked and executed by the physical robot. To our knowledge, this work represents a significant step toward general real-world use by providing a complete pipeline for automatic, semantic-aware, co-speech gesture generation and synchronized real-time physical deployment on a humanoid robot.
\end{abstract}

\maketitle

\section{Introduction}

Enabling robots to express themselves naturally, akin to human interaction, remains a long-standing aspiration within the field of robotics. Human-like behaviors, including non-verbal cues such as gestures, are crucial for effective communication and for creating relatable and engaging virtual or physical characters. While rhythmic movements (beat gestures) are relatively common and easily synthesized, gestures that convey specific semantic meaning (semantic gestures) are sparse in datasets, making their accurate generation a persistent challenge for data-driven systems. Furthermore, converting these complex, human-centric motions into executable commands for a physical humanoid robot, while maintaining balance and agility in real-time environments, introduces additional significant challenges related to the embodiment gap and robust physical control.

This work addresses the full pipeline, from semantic speech understanding to real-time physical deployment of co-speech gestures on the Unitree G1 humanoid robot. Our methodology combines recent advancements in semantics-aware gesture synthesis with a dedicated imitation learning control framework designed for robust humanoid operation.

The primary contributions and sequence of our work are outlined as follows:
\begin{enumerate}
    \item \textbf{Data Acquisition and Retargeting:} We first collect a speech and gesture alignment dataset, where speech is stored as WAV audio and gesture data as BVH motion files. We employ the General Motion Retargeting (GMR) method to accurately transfer these BVH motions from the human skeleton to the specific kinematic and morphological structure of the Unitree G1 robot.
    \item \textbf{Semantic Gesture Synthesis Pipeline:} Drawing inspiration from recent frameworks, we train a motion encoder and decoder (Residual VQ-VAE) to learn a discrete motion space. We then train an autoregressive Motion-GPT model, conditioned on audio input, to automatically generate semantically coherent and rhythmically matched gestures, represented as motion tokens.
    \item \textbf{Imitation Learning Control Training:} Using the retargeted G1 motion data as high-fidelity reference trajectories, we train a robust humanoid control algorithm, the MotionTracker, based on imitation learning (IL) principles. This controller is designed to execute diverse and dynamic motions efficiently.
    \item \textbf{Real-Time Deployment:} During inference, the system generates the corresponding speech (via a Text-to-Speech tool) and simultaneously uses the Motion-GPT to generate matching gesture tokens. These tokens are decoded into reference motion, fed to the control policy, which then drives the robot to perform the gestures in real-time, synchronized with the spoken audio.
\end{enumerate}

\section{Related Work}

\subsection{Gesture Generation}
Early efforts in co-speech gesture synthesis primarily relied on rule-based approaches, which utilized linguistic mappings to convert speech into sequences of predefined gesture clips. While offering interpretability, these methods were labor-intensive and resulted in non-realistic movements. The field transitioned toward data-driven approaches to reduce manual effort and enhance motion quality. Statistical models were initially employed to extract mappings from data, but eventually, deep neural networks became the dominant methodology, enabling end-to-end training directly from raw gesture data.

Deep learning methods often utilize models like MLPs, CNNs, RNNs, and Transformers, frequently achieving success in generating beat gestures that align rhythmically with speech. However, many systems struggle to produce semantic gestures (gestures reflecting the meaning of spoken words) due to their inherent sparsity, treating them as noise in training data. To address this, techniques like clustering and resampling have been attempted to improve data balance. More recently, models have incorporated retrieval augmentation, such as using large language models (LLMs) to retrieve relevant semantic gesture candidates from external libraries based on textual context, thereby enhancing the generation of complex, sparse movements \cite{Zhang2024SemanticGesture}. Zero-shot approaches have also emerged, enabling gesture generation with style control by example without requiring explicit style-specific training data \cite{ghorbani2022zeroeggs}. Generative models, including VAEs, VQ-VAEs, and diffusion models, are widely used to mitigate the mean collapse problem and generate varied outputs.

\subsection{Humanoid Motion Control}
The goal of humanoid motion control, particularly motion tracking, is to reproduce human motion sequences on a robot platform to achieve expressive and anthropomorphic movement behaviors \cite{chen2025gmt, ji2024exbody2}. This task is complicated by the cross-embodiment gap, physical actuator limits, and the sim-to-real transfer challenge \cite{he2025asap}. Existing humanoid motion trackers demonstrate varying limitations in motion tracking capability and resilience to real-world disturbances \cite{he2024omnih2o}. While some approaches focus on quasi-static tasks or short, highly dynamic clips, foundational motion trackers must handle diverse, highly dynamic, and contact-rich movements across extensive datasets like AMASS and LAFAN1.

A critical aspect of robust deployment is online dynamics adaptation, enabling the robot to handle real-world disturbances such as changing terrains, external forces, and physical property variations (e.g., payloads) \cite{zhang2025track}. Simple domain randomization is often insufficient against large dynamic variance. More advanced approaches utilize interaction history to estimate environment dynamics and adapt behavior, often employing two-stage frameworks to decouple the learning of basic motion execution capability from dynamics adaptability. For instance, the Any2Track framework proposes the AnyTracker for general motion execution and the AnyAdapter for history-informed dynamics adaptation \cite{zhang2025track}.

\section{Methodology}

Our system integrates three primary components: motion retargeting, automatic gesture generation, and robust imitation learning control.

\subsection{System Overview}

\begin{figure*}
    \centering
    \includegraphics[width=\textwidth]{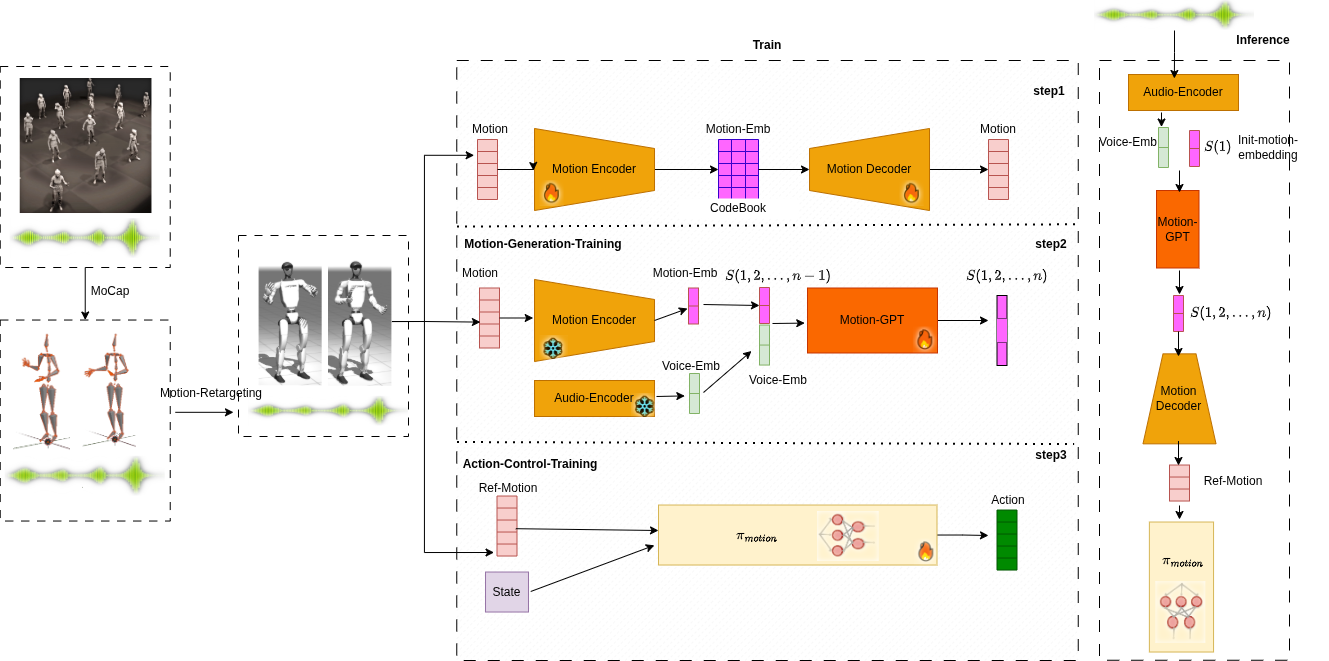}
    \caption{System Overview: Training and Inference Pipeline.}
    \label{fig:system_overview}
\end{figure*}

As illustrated in Figure \ref{fig:system_overview}, our system architecture consists of a training phase divided into three stages and a final inference pipeline.

The training phases are:
\begin{enumerate}
    \item \textbf{Motion Codebook Training (step 1):} A Residual VQ-VAE learns a hierarchical discrete latent space for motion, mapping motion sequences to compressed tokens.
    \item \textbf{Motion Generation Training (step 2):} An autoregressive Motion-GPT model is trained to predict future motion tokens conditioned on preceding motion tokens and synchronized audio features.
    \item \textbf{Action Control Training (step 3):} A specialized imitation learning policy, $\pi_{motion}$, is trained using reinforcement learning (RL) to track high-fidelity reference motions (Ref-Motion) generated for the target robot.
\end{enumerate}

During inference, audio is processed by the Audio-Encoder and fed into the Motion-GPT, which generates a sequence of motion tokens. These tokens are decoded into a Ref-Motion, which in turn guides the $\pi_{motion}$ controller to output real-time actions for the robot.

\subsection{Motion Retargeting}

Motion retargeting is essential to address the embodiment gap between general human motion data (e.g., BVH format) and the specific morphology of the humanoid robot (Unitree G1) \cite{joao2025gmr}. We leverage the principles of General Motion Retargeting (GMR) to create high-quality reference motions, as artifacts like ground penetration, self-intersection, or abrupt velocity spikes introduced during poor retargeting can severely degrade the subsequent policy training's robustness \cite{joao2025gmr}.

Our process involves:
\begin{enumerate}
    \item \textbf{Key Body Matching and Alignment:} We first define the correspondence between the human and robot key bodies and align their rest poses in Cartesian space. This initialization step helps mitigate artifacts such as the "toed-in" effect.
    \item \textbf{Non-Uniform Local Scaling:} We introduce a scaling procedure to adjust the source motion's proportions to match the robot. Crucially, we use a uniform scaling factor applied only to the root translation, which is necessary to avoid introducing foot sliding artifacts.
    \item \textbf{Two-Stage Kinematic Optimization:} The final robot motion is determined using a two-stage optimization process based on a differential inverse kinematics (IK) solver. The first stage focuses on minimizing body orientation errors and end-effector position errors, providing a stable initialization. The second stage then fine-tunes the solution by including position constraints for all key bodies.
\end{enumerate}
The resulting G1 motion data includes the root position, root linear velocity, root angular velocity, and the joint angles for the robot's 29 degrees of freedom (DoFs).

\subsection{Automatic Gesture Generation}

The core of generating semantically appropriate co-speech gestures relies on learning a discrete motion space and an autoregressive model conditioned on speech, inspired by recent advances in gesture synthesis.

\subsubsection{Motion Autoencoder (Residual VQ-VAE)}
We utilize a Residual Vector Quantized Variational Autoencoder (RVQ-VAE) to learn a hierarchical categorical space that represents motion as discrete tokens. This is critical because standard VQ-VAEs often lack the representational capacity to reconstruct complex movements, especially fine details like finger articulation. The RVQ-VAE enhances capacity by:
\begin{enumerate}
    \item Separating motion representation into body and hand parts, modeling them independently to handle movement complexity efficiently.
    \item Employing multiple residual quantization layers in a hierarchical architecture, iteratively modeling motion features and exponentially expanding the model's expressive capability without unstable training issues like code collapse.
\end{enumerate}

\subsubsection{Autoregressive Motion-GPT Model}
The gesture generative model ($G$) is based on the GPT-2 architecture, designed to generate rhythm-matched gestures. This model predicts the future discrete gesture tokens (for both hand and body) in an autoregressive fashion, conditioned on preceding motion tokens and synchronized audio features. The audio features incorporate various cues like MFCC, delta, chromagram, onset, and tempogram, extracted from the audio signal.

Furthermore, to ensure the synthesis of semantically rich, yet sparse, gestures, the framework incorporates a large language model (LLM)-based retrieval mechanism. This LLM analyzes the speech transcript context to efficiently retrieve appropriate semantic gesture candidates from a high-quality gesture library (SeG dataset). These retrieved semantic gestures are then fused with the rhythmically generated motion at the latent space level via a semantics-aware alignment mechanism, ensuring the final animation is both natural and contextually meaningful \cite{Zhang2024SemanticGesture}.

\subsection{Imitation Learning Control Policy}

To enable the robot to physically execute the generated reference motions, we utilize the MotionTracker, developed within the two-stage reinforcement learning (RL) framework. MotionTracker serves as a general motion tracker designed to track diverse, highly dynamic, and contact-rich human motions, minimizing the tracking error between the robot's state and the target motion goals \cite{zhang2025track}.

Recent advances in humanoid whole-body control have demonstrated promising results for tracking diverse and dynamic human motions \cite{chen2025gmt, ji2024exbody2}. These approaches often focus on developing general motion tracking frameworks that can handle complex whole-body coordination while maintaining balance and stability \cite{he2024omnih2o}. Additionally, techniques that align simulation and real-world physics have shown significant improvements in transferring learned skills from simulation to physical robots \cite{he2025asap}.

To overcome the optimization difficulties associated with the robot's high degrees of freedom (DoFs) and the diversity of motion data, MotionTracker rather than predicting absolute PD targets, the policy predicts residual PD offsets relative to the reference motion. The final action, which is mapped using a $\tanh$ function, is scaled by empirically designed hyperparameters for each joint. This canonicalization allows the policy to predict a compact, multi-joint action distribution effectively.
 
This methodology ensures that the learned policy maintains high expressiveness and motion tracking fidelity when executing the gestures synthesized by the Motion-GPT module.

\section{Experiments}

\subsection{Dataset}
Our system is trained and evaluated using high-quality multimodal datasets. For gesture synthesis, we draw upon resources including the ZEGGS dataset and the BEAT dataset. The ZEGGS dataset, for example, features two hours of full-body motion capture and audio from a single English-speaking female actor, covering 19 distinct motion styles.

\subsection{Motion Autoencoder Validation}

The purpose of this experiment is to validate the reconstruction capability of the Residual VQ-VAE (RVQ-VAE) motion autoencoder. We verify that the model can accurately encode motion sequences into discrete tokens and reconstruct them with minimal error, ensuring the foundation of the motion generation pipeline is solid.

\begin{figure}[h]
    \centering
    \includegraphics[width=0.95\linewidth]{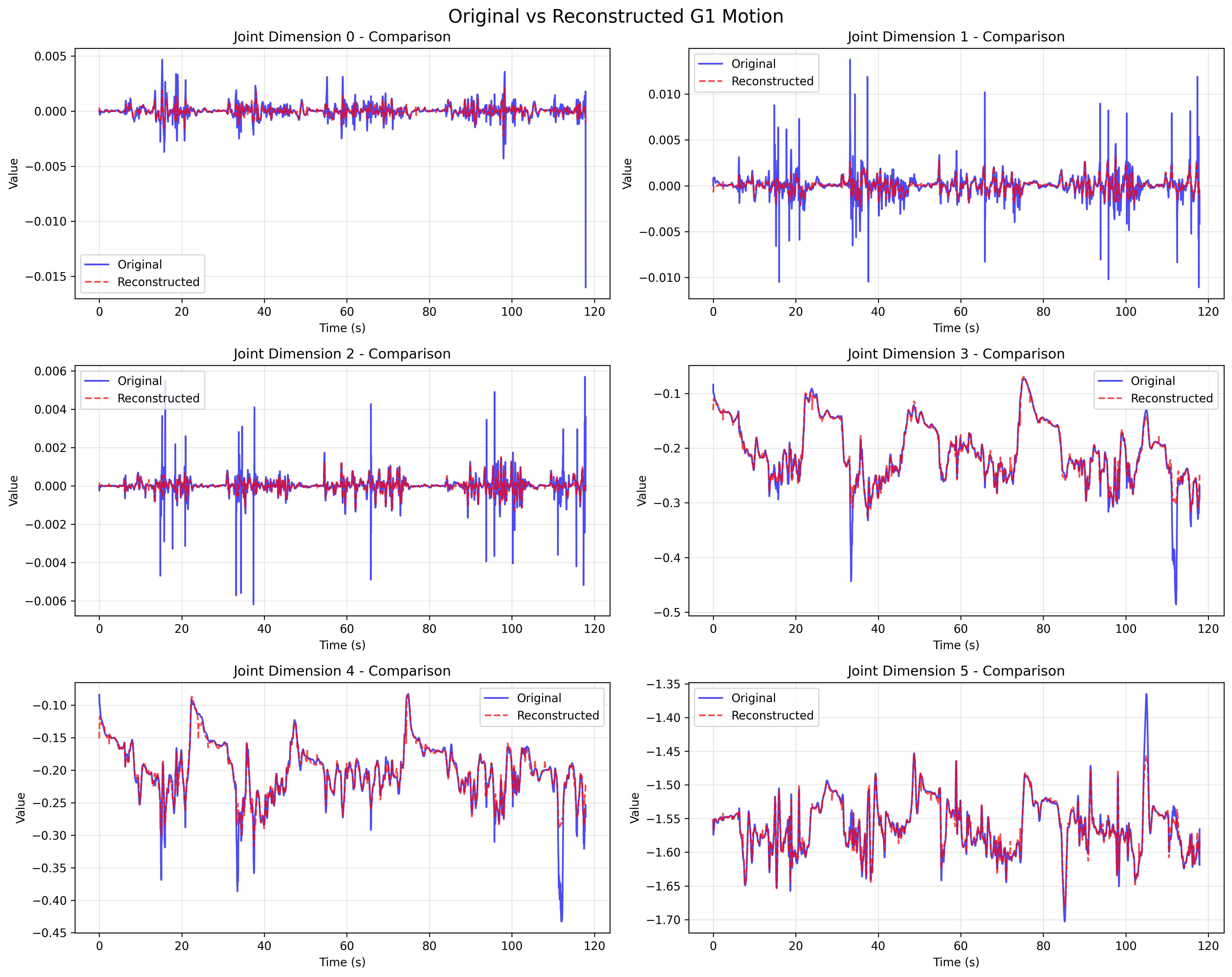}
    \caption{Comparison of Original vs. Reconstructed G1 Motion (Joint Dimensions 0-5). The reconstructed motion (red dashed line) closely follows the original motion (blue solid line) over time.}
    \label{fig:rvq_comparison}
\end{figure}

Figure \ref{fig:rvq_comparison} displays the comparison between the original (Ground Truth, blue solid line) and reconstructed (Reconstructed, red dashed line) motion values for six randomly selected joint dimensions over a 120-second sequence. The reconstructed joint angles exhibit a close fit to the ground truth data, demonstrating that the RVQ-VAE effectively encodes motion information, allowing for accurate reconstruction and minimal information loss during the tokenization process.

\subsection{Gesture Generation Quality}

This experiment evaluates the quality and accuracy of the gestures generated by the Motion-GPT model conditioned on audio input. We assess how well the generated movements align with the corresponding human-recorded ground truth gestures.

\begin{figure}[h]
    \centering
    \includegraphics[width=0.95\linewidth]{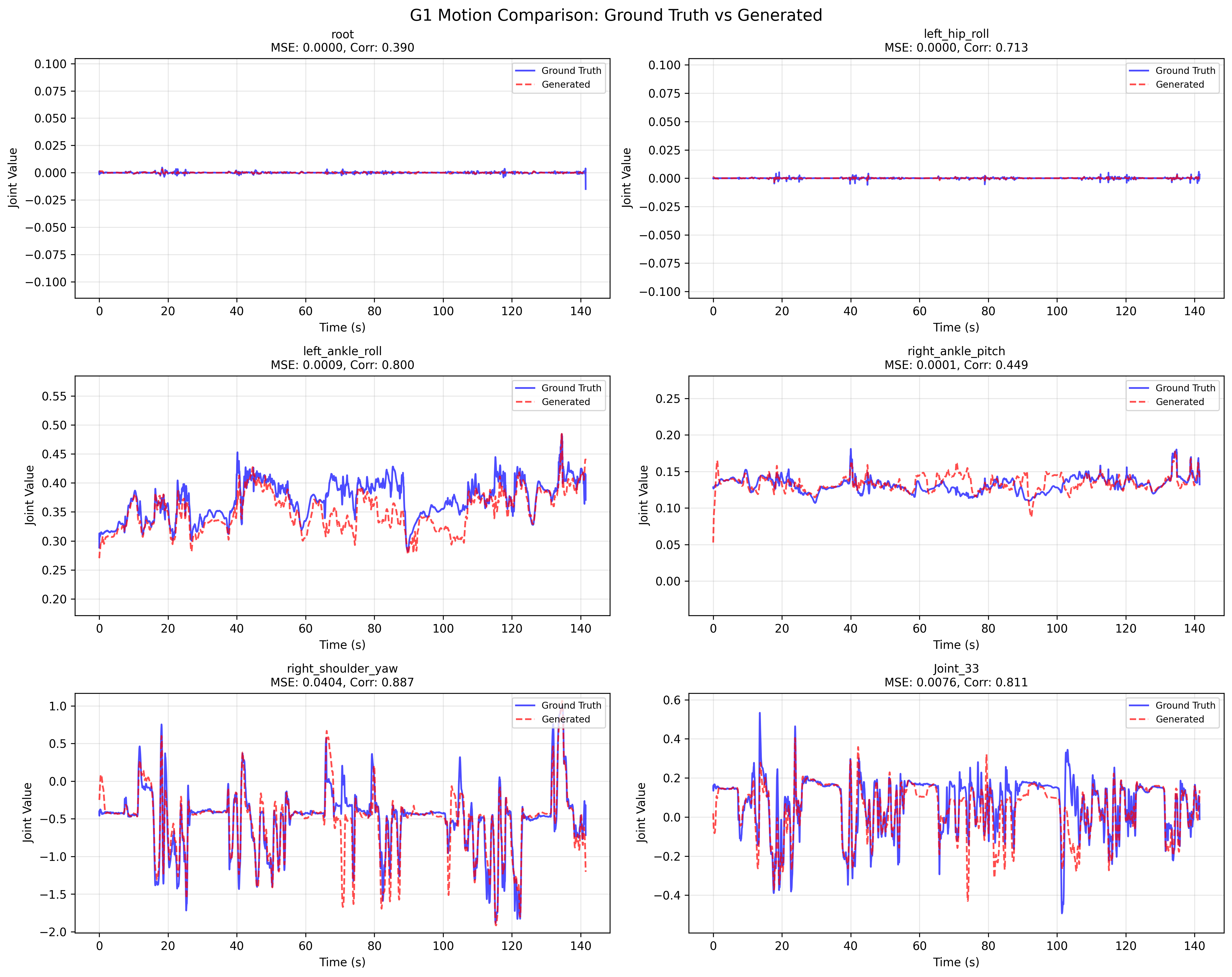}
    \caption{G1 Motion Comparison: Ground Truth vs. Generated (Selected Joints). The generated motion shows a strong correlation and low MSE with the Ground Truth data.}
    \label{fig:gesture_comparison}
\end{figure}

Figure \ref{fig:gesture_comparison} compares the Ground Truth motion (blue solid line) against the Generated motion (red dashed line) for a selection of joints over time. The visual evidence confirms that the motion data generated based on the audio input exhibits a good fit with the original reference gesture data in terms of joint angles.

The quantitative joint-wise error metrics (Root Mean Square Error, RMSE) further confirm the high quality of the generated motion (Table \ref{tab:gesture_generation_rmse}). Notably, errors for root position across all three axes are very low (e.g., root\_pos\_x: $0.000592$ and root\_pos\_y: $0.000783$).

\begin{table}[h]
\caption{Joint-wise Error Metrics (RMSE) for Gesture Generation Quality}
\centering
\begin{tabular}{|l|l|}
\hline
\textbf{Joint} & \textbf{RMSE} \\ \hline
root\_pos\_x & 0.000592 \\
root\_pos\_y & 0.000783 \\
root\_pos\_z & 0.000329 \\
root\_vel\_x & 0.025074 \\
root\_vel\_y & 0.019808 \\
root\_vel\_z & 0.043517 \\ \hline
left\_hip\_pitch\_joint & 0.029306 \\
left\_hip\_roll\_joint & 0.014403 \\
left\_hip\_yaw\_joint & 0.183834 \\
left\_knee\_joint & 0.034635 \\
left\_ankle\_pitch\_joint & 0.019814 \\
left\_ankle\_roll\_joint & 0.011305 \\ \hline
right\_hip\_pitch\_joint & 0.020059 \\
right\_hip\_roll\_joint & 0.013712 \\
right\_hip\_yaw\_joint & 0.176344 \\
right\_knee\_joint & 0.038665 \\
right\_ankle\_pitch\_joint & 0.017804 \\
right\_ankle\_roll\_joint & 0.009490 \\ \hline
waist\_yaw\_joint & 0.034864 \\
waist\_roll\_joint & 0.026776 \\
waist\_pitch\_joint & 0.020413 \\ \hline
left\_shoulder\_pitch\_joint & 0.078946 \\
left\_shoulder\_roll\_joint & 0.044724 \\
left\_shoulder\_yaw\_joint & 0.159468 \\
left\_elbow\_joint & 0.183838 \\
left\_wrist\_roll\_joint & 0.200901 \\
left\_wrist\_pitch\_joint & 0.081273 \\
left\_wrist\_yaw\_joint & 0.169707 \\ \hline
right\_shoulder\_pitch\_joint & 0.064324 \\
right\_shoulder\_roll\_joint & 0.040821 \\
right\_shoulder\_yaw\_joint & 0.159764 \\
right\_elbow\_joint & 0.196378 \\
right\_wrist\_roll\_joint & 0.226248 \\
right\_wrist\_pitch\_joint & 0.086990 \\
right\_wrist\_yaw\_joint & 0.180439 \\ \hline
\end{tabular}
\label{tab:gesture_generation_rmse}
\end{table}

\subsection{Imitation Learning Control Policy}

To validate the motion control policy's effectiveness ($\pi_{motion}$), we compare the simulated robot’s tracking performance against the reference motion.

\subsubsection{Simulation Tracking Accuracy}
Table \ref{tab:imitation_learning_rmse} shows the RMSE errors between the robot's state (joint angles and velocities) and the reference motion in the simulation environment. The errors are generally low across all DoFs, indicating that the imitation learning algorithm successfully trained the robot to accurately follow the highly dynamic reference motions generated by the gesture synthesis pipeline. For example, lower body movements show low RMSE, such as the right hip roll joint (0.0064).

\begin{table}[h]
\caption{Joint-wise Performance: Imitation Learning RMSE Error}
\centering
\begin{tabular}{|l|l|}
\hline
\textbf{Joint} & \textbf{RMSE} \\ \hline
root\_pos\_x & 0.0109 \\
root\_pos\_y & 0.0139 \\
root\_pos\_z & 0.0050 \\
root\_vel\_x & 0.0261 \\
root\_vel\_y & 0.0510 \\
root\_vel\_z & 0.0228 \\ \hline
left\_hip\_pitch\_joint & 0.0504 \\
left\_hip\_roll\_joint & 0.0127 \\
left\_hip\_yaw\_joint & 0.0394 \\
left\_knee\_joint & 0.0644 \\
left\_ankle\_pitch\_joint & 0.0594 \\
left\_ankle\_roll\_joint & 0.0760 \\ \hline
right\_hip\_pitch\_joint & 0.0532 \\
right\_hip\_roll\_joint & 0.0064 \\
right\_hip\_yaw\_joint & 0.0975 \\
right\_knee\_joint & 0.0879 \\
right\_ankle\_pitch\_joint & 0.0297 \\
right\_ankle\_roll\_joint & 0.0508 \\ \hline
waist\_yaw\_joint & 0.0465 \\
waist\_roll\_joint & 0.0101 \\
waist\_pitch\_joint & 0.0486 \\ \hline
left\_shoulder\_pitch\_joint & 0.0508 \\
left\_shoulder\_roll\_joint & 0.0226 \\
left\_shoulder\_yaw\_joint & 0.0445 \\
left\_elbow\_joint & 0.0612 \\
left\_wrist\_roll\_joint & 0.0864 \\
left\_wrist\_pitch\_joint & 0.0443 \\
left\_wrist\_yaw\_joint & 0.0841 \\ \hline
right\_shoulder\_pitch\_joint & 0.0352 \\
right\_shoulder\_roll\_joint & 0.0382 \\
right\_shoulder\_yaw\_joint & 0.0569 \\
right\_elbow\_joint & 0.0367 \\
right\_wrist\_roll\_joint & 0.0910 \\
right\_wrist\_pitch\_joint & 0.0532 \\
right\_wrist\_yaw\_joint & 0.0765 \\ \hline
\end{tabular}
\label{tab:imitation_learning_rmse}
\end{table}

\subsubsection{Real-World Deployment}
\begin{figure}[h]
    \centering
    \includegraphics[width=0.95\linewidth]{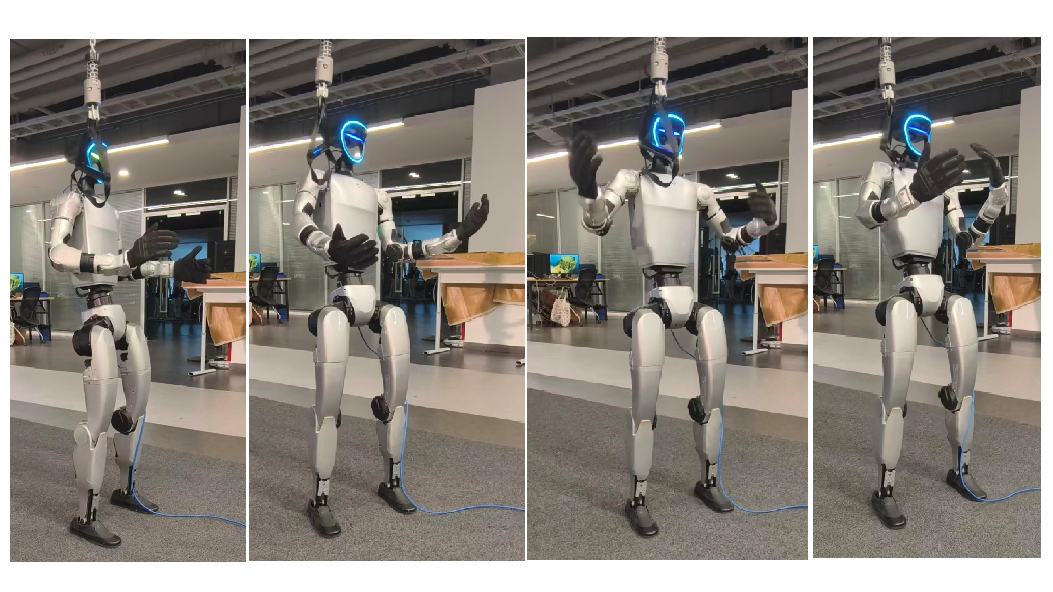}
    \caption{Real-World Deployment Demonstration on Unitree G1. The robot executes generated co-speech gestures while maintaining balance.}
    \label{fig:real_deploy}
\end{figure}

To validate the complete end-to-end performance, the motion control policy ($\pi_{motion}$) was deployed on the physical Unitree G1 humanoid robot. Figure \ref{fig:real_deploy} provides visual evidence of the robot executing the synthesized gestures.

The deployment demonstrates that the model can successfully take speech input, automatically generate matching semantic gestures, and then drive the robot to execute these gestures in real-time. Crucially, the robot is able to maintain its balance while performing the hand and body movements, validating the robustness of the combined gesture generation and imitation control system for practical, real-world use.

\section{Conclusion}

We successfully developed and implemented a novel end-to-end system for Semantic Co-Speech Gesture Synthesis and Real-Time Control for Humanoid Robots. Our framework effectively bridges the gap between high-level communicative intent embedded in speech and low-level physical control required for dynamic robotic execution. By combining the strengths of the GMR motion retargeting method, a powerful LLM-augmented Motion-GPT model for semantic gesture synthesis, and the robust MotionTracker policy for imitation learning control, we demonstrated that humanoid robots can be driven to perform expressive and semantically rich gestures synchronized with audio in real time. Quantitative evaluations confirmed the high fidelity of motion encoding, gesture generation, and control tracking accuracy, while real-world deployment validated the stability and effectiveness of the integrated system.

Future work should explore enhancing the LLM's understanding of speech rhythm and prosody to refine the timing and frequency of semantic gesture retrieval. Furthermore, investigating more complex merging strategies beyond simple stroke phase alignment could better preserve motion details and naturalness. Finally, extending this framework to multi-party conversational scenarios remains a promising direction for creating truly interactive and anthropomorphic humanoid agents.

\bibliographystyle{plain}
\bibliography{ref}

\end{document}